%% file: main.tex
\documentclass[10pt,twocolumn,letterpaper]{article}

\usepackage{cvpr}              

\input{preamble}

\usepackage{siunitx}     
\definecolor{cvprblue}{rgb}{0.21,0.49,0.74}
\usepackage[pagebackref,breaklinks,colorlinks,allcolors=cvprblue]{hyperref}
\usepackage{multirow}
\usepackage{booktabs}
\usepackage{makecell}
\usepackage{ulem}
\usepackage{cuted}

\usepackage[table]{xcolor}

\def\paperID{}
\def\confName{CVPR}
\def\confYear{2026}

\title{Decomposing Subject-Driven Image Generation \\ via Intermediate Structural Prediction}

\author{
Hanzhong Guo \qquad
Yizhou Yu\\[0.5em]
School of Computing and Data Science, The University of Hong Kong \\
{\tt\small hanzhong@connect.hku.hk, yizhouy@acm.org}
}
\begin{document}
\maketitle

\begin{strip}
\centering
\vspace{-1.5cm}
\includegraphics[width=\textwidth]{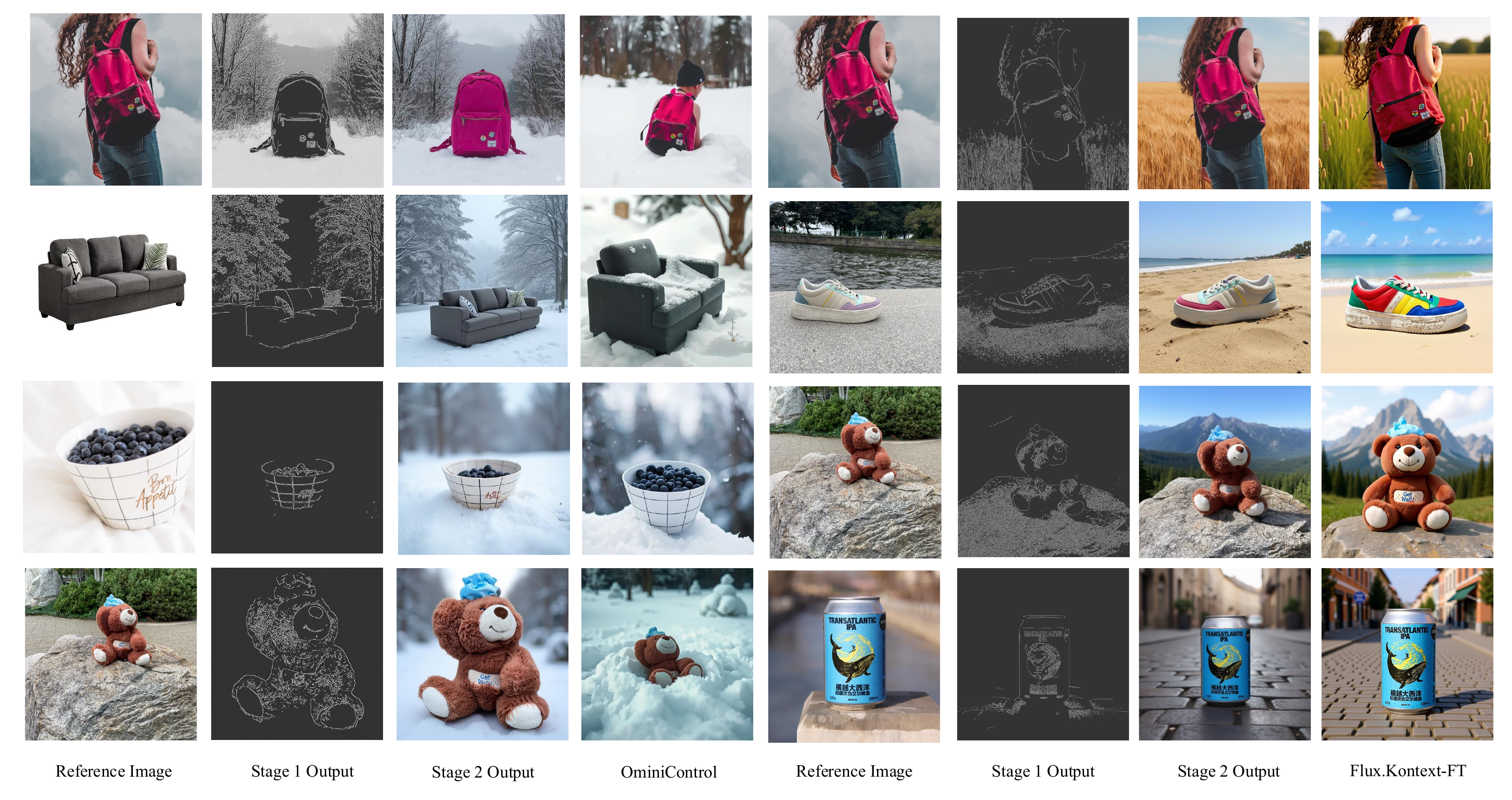}
\captionof{figure}{
    \textbf{High-Fidelity Subject-Driven Generation through Structural Decomposition.} Our method excels at preserving the identity of subjects, especially those with high-frequency details like text and patterns. 
    \textbf{(Left Panel)} We demonstrate our primary approach on a frozen FLUX.1-dev backbone. Given a reference image, our method first predicts a target Canny map (\textit{Stage 1 Output}) that captures the desired structural changes. Then, it renders the final image (\textit{Stage 2 Output}) conditioned on this structure and the source appearance. Compared to a strong baseline like OminiControl~\cite{tan2024ominicontrol}, which often distorts or loses fine details (e.g., the backpack's pattern, the teddy bear's hat), our method maintains superior identity fidelity.
    \textbf{(Right Panel)} We showcase the versatility of our two-stage paradigm by adapting it to a different backbone, FLUX.Kontext, via LoRA fine-tuning (Flux.Kontext-FT). The framework again successfully preserves intricate details, such as the text on the beverage can, demonstrating its robustness and broad applicability.
}
\label{fig:teaser}
\end{strip}

\input{sec/0_abstract}
\input{sec/1_intro}

\input{sec/2_related}

\input{sec/3_method}

\input{sec/4_exps}

\input{sec/5_conclu}
\section*{Acknowledgments}
This work was supported by Hong Kong Research Grants Council under NSFC/RGC Collaborative Research Scheme (Grant CRS\_HKU703/24).

\clearpage
{
    \small
    \normalem
    \bibliographystyle{ieeenat_fullname}
    \bibliography{main}
}
\input{sec/X_supp}

\end{document}

%% file: preamble.tex
\usepackage[dvipsnames]{xcolor}
\usepackage{bbding}

\usepackage{times}
\usepackage{epsfig}
\usepackage{graphicx}
\usepackage{amsmath}
\usepackage{amssymb}
\usepackage{xcolor}
\usepackage{caption}



%% file: sec/0_abstract.tex
\begin{abstract}
Subject-driven text-to-image generation still struggles to preserve high-frequency identity details such as logos, patterns, and text. Existing methods typically operate directly in RGB space, which often leads to detail degradation under substantial edits. We propose a two-stage framework that decouples structure from appearance by first predicting a Canny map and then rendering the final image conditioned on both the source appearance and the predicted structure. To improve text handling, we further introduce a fully automatic pipeline that constructs a 100k-pair text-aware dataset with cross-view textual consistency. Experiments, including GPT-4.1-based evaluation and a knowledge distillation study, show clear gains over selected baselines and suggest that intermediate structural prediction is an effective route for high-fidelity subject-driven generation. Our dataset and code will be made publicly available.
\end{abstract}

%% file: sec/1_intro.tex
\section{Introduction}
\label{sec:intro}
The advent of large-scale text-to-image (T2I) models, particularly Diffusion Transformers (DiT)~\cite{peebles2023scalable} and latent diffusion models~\cite{rombach2022high}, has revolutionized digital content creation. Building on this foundation, subject-driven generation has emerged as a highly sought-after application, aiming to generate novel renditions of a specific, user-provided subject in new scenes or styles. However, despite rapid progress, a fundamental challenge persists: the trade-off between editability and identity fidelity. As illustrated in Figure~\ref{fig:teaser}, while existing methods can place a subject in a new context, they often fail to preserve its defining, high-frequency details. A logo on a can becomes a blurry smudge; the unique pattern on a backpack is averaged into a generic texture. This degradation of identity undermines the very purpose of subject-driven generation.
This failure stems from a core methodological flaw. Current leading approaches, from full fine-tuning methods like DreamBooth~\cite{ruiz2022dreambooth} to more efficient adapter-based models like IP-Adapter~\cite{ye2023ip} and OminiControl~\cite{tan2024ominicontrol}, attempt to learn a direct, single-step mapping from source inputs to the final RGB image. This forces the model to solve an ill-posed and entangled optimization problem: it must simultaneously process global compositional changes (e.g., altering pose and background) and preserve local, high-frequency details (e.g., texture and text). In this entangled space, the semantic priors of the base model can overwhelm the subtle identity features of the subject, leading to the loss of detail shown by the OminiControl baseline in Figure~\ref{fig:teaser}.
To address this, we propose a solution that reframes the learning problem itself. Our core contribution, illustrated in the central columns of Figure~\ref{fig:teaser}, is a two-stage framework that decouples the generation process into two more manageable sub-problems: \textbf{structure prediction} and \textbf{appearance rendering}. We observe that Canny edge maps~\cite{canny1986computational} provide a sparse yet powerful representation of an object's high-frequency structural information. Therefore, our method first predicts a target Canny map, effectively creating a structural blueprint that dictates the subject's form and fine details in the new context. Only then does a second stage render the final image, guided by this explicit structural information and the original subject's appearance.
This decomposition does more than preserve details; it offers a practical methodology for subject-driven generation. By breaking down the complex direct-to-RGB task, we provide the model with a form of curriculum. The first stage solves a simpler, lower-dimensional problem about how the key structural lines go.
The second stage then tackles a more constrained synthesis problem about how render given this structure and the source appearance.
This structured approach simplifies the learning objective at each step and leads to more stable training.
The versatility of this paradigm is another key advantage. As shown on the right of Figure~\ref{fig:teaser} and further examined in our knowledge distillation experiment (Sec.~\ref{sec:exps}), our framework is not tied to a single architecture. It can be adapted to backbones such as FLUX.Kontext through LoRA fine-tuning, suggesting that intermediate structural prediction is a promising technical route beyond the specific FLUX.1-dev setting used in our main experiments. At the same time, we regard broader validation on larger and more recent backbones as an important limitation and future direction.
Finally, to help the model handle one of the most challenging high-frequency cases, text rendering, we construct a new large-scale (100k) dataset, `TextingSubject100k`, using a fully automated pipeline. This dataset directly addresses a critical data scarcity issue in text-aware subject customization. By combining our two-stage framework with this specialized data, and by adopting a minimal control architecture, our method achieves strong identity preservation while remaining lightweight.
Our contributions are threefold:
\begin{itemize}
    \item A novel two-stage framework that decouples structure and appearance, offering a practical methodological route for high-fidelity subject-driven generation.
    \item A new data generation pipeline and a resulting large-scale (100k) dataset, `TextingSubject100k`, specifically designed for the challenging task of text-on-object customization.
    \item Extensive validation, including a GPT-4.1-based evaluation and a knowledge distillation experiment, showing clear gains over selected baselines and supporting the effectiveness of the proposed decomposition.
\end{itemize}

%% file: sec/2_related.tex
\section{Related Works}
\label{sec:related}

Our work is positioned at the intersection of three active research areas: subject-driven image generation, which focuses on personalization; controllable image synthesis, which provides explicit spatial guidance; and the specialized challenge of text rendering in generative models.

\subsection{Subject-Driven Image Generation}
Subject-driven generation aims to personalize a text-to-image (T2I) model with a novel concept provided by a user, typically through one or more reference images. Early and highly influential methods relied on fine-tuning parts of the T2I model. DreamBooth~\cite{ruiz2022dreambooth}, for instance, fine-tunes the entire diffusion model (U-Net) with a unique text identifier, achieving remarkable identity preservation but at a significant computational and storage cost for each new subject. To improve efficiency, a line of work has focused on lightweight alternatives. Textual Inversion~\cite{gal2022image} keeps the main model frozen and instead optimizes a new "word" embedding in the text encoder's space to represent the subject. CustomDiffusion~\cite{kumari2023customdiffusion} offers a middle ground by fine-tuning only the efficient cross-attention layers and a small portion of the U-Net.

More recently, the field has shifted towards even more efficient, tuning-free paradigms that leverage pretrained encoders. Methods like IP-Adapter~\cite{ye2023ip} introduce a decoupled cross-attention mechanism, allowing an adapter to be trained on image features. This enables strong performance without fine-tuning the base model for each new subject. Following this trend, other works like PhotoMaker~\cite{li2024photomaker} and InstantID~\cite{wang2024instantid} have pushed the boundaries of identity preservation for human faces by using more powerful identity encoders and fusing features more effectively. Related lines of work also explore identity-sensitive video synthesis~\cite{guo2024expressiveportrait} and defenses against misuse of personalization~\cite{guo2024real}. While these methods excel at capturing the global appearance and low-frequency characteristics of a subject, they share a common limitation: a struggle to faithfully reproduce high-frequency details. As we argue in our introduction, this is because they often encode the subject into a holistic feature vector, which tends to average out fine patterns, text, and intricate textures. Our work fundamentally diverges from this paradigm. Instead of seeking a better holistic feature representation, we propose that true fidelity requires explicitly modeling the subject's structure. By predicting a Canny map as an intermediate step, we shift the problem from direct feature injection to structure-guided synthesis.

\subsection{Controllable Image Generation}
To provide users with more precise and explicit control over the generation process, a parallel line of work has focused on incorporating spatial guidance~\cite{guo2026leveraging}. ControlNet~\cite{zhang2023adding} and T2I-Adapter~\cite{mou2023t2i} are seminal works in this domain. They introduced a groundbreaking mechanism to inject spatial conditions—such as Canny edges, depth maps, or human poses—into a frozen T2I model backbone. This is typically achieved by creating a trainable copy of the model's encoding layers, which processes the control signal and feeds the guidance into the main model via simple addition. This allows for robust spatial control without compromising the generative capabilities of the pretrained model.

With the rise of more powerful Diffusion Transformers (DiT)~\cite{peebles2023scalable} and efficient diffusion backbones~\cite{fu2025lamambadiff}, this control paradigm has also evolved. Neural image transformation itself has a long history in graphics, including learned global photo adjustment~\cite{yan2016automatic}. OminiControl~\cite{tan2024ominicontrol} proposed a minimal and universal control mechanism specifically for transformer-based diffusion models. Instead of copying large parts of the network, it achieves robust control by inserting very small, lightweight trainable blocks, making it extremely parameter-efficient. Our work is inspired by and adopts this minimal control paradigm for our architecture. However, a critical distinction exists. Standard controllable generation methods assume the control signal (e.g., a Canny map) is provided by the user. Our task is different: we do not have the target Canny map. Our novelty lies in bridging subject-driven generation with controllable synthesis: we first \textit{predict} the necessary control signal based on the source subject and the text prompt. Thus, our framework does not just \textit{use} control; it \textit{generates} the control as a core part of its reasoning process, making it a more intelligent and automated form of controlled generation.

\subsection{Text Rendering in Generative Models}
Rendering legible, coherent, and aesthetically pleasing text in images is a known grand challenge for standard diffusion models. These models, trained primarily on natural images, struggle with the precise, compositional, and stroke-level accuracy required for typography. To address this, a dedicated subfield has emerged with specialized solutions. TextDiffuser~\cite{chen2023textdiffuser} proposed a two-stage framework involving character-aware attention and an explicit layout planning stage to guide text placement. More recent methods like GlyphControl~\cite{zeng2024glyphcontrol} adopt the ControlNet paradigm, taking a binary glyph image (a black-and-white image of the desired text) as an additional condition to guide the diffusion process, significantly improving text quality.

These methods are highly effective for their intended task: generating images with text from scratch. However, our task is different and arguably more complex. We aim to customize a \textit{specific existing object} to include, modify, or retain text, all while preserving the object's unique identity, often across different views or poses. This introduces a challenging three-way constraint between (1) the source subject's identity, (2) the desired textual content and style, and (3) the overall scene described by the prompt. Existing text rendering models are not designed to handle the subject identity constraint. This is precisely the gap our work aims to fill. By creating the `TextingSubject100k` dataset, we provide the first large-scale resource to learn this complex relationship. Our two-stage model is then specifically designed to leverage this data, using the predicted structure to ensure that the rendered text is not only legible but also correctly positioned and warped according to the subject's geometry, thereby maintaining object integrity.

Beyond architecture design, data-centric improvements through synthetic data or targeted augmentation have also shown value in adjacent settings, e.g., DreamDA~\cite{fu2024dreamda} and CarveMix~\cite{zhang2023carvemix}. Our `TextingSubject100k` pipeline follows this broader philosophy, but is tailored specifically to subject-consistent, text-bearing object customization.

%% file: sec/3_method.tex
\section{Method}
\label{sec:method}
\begin{figure*}[t]
  \centering
  \includegraphics[width=\linewidth]{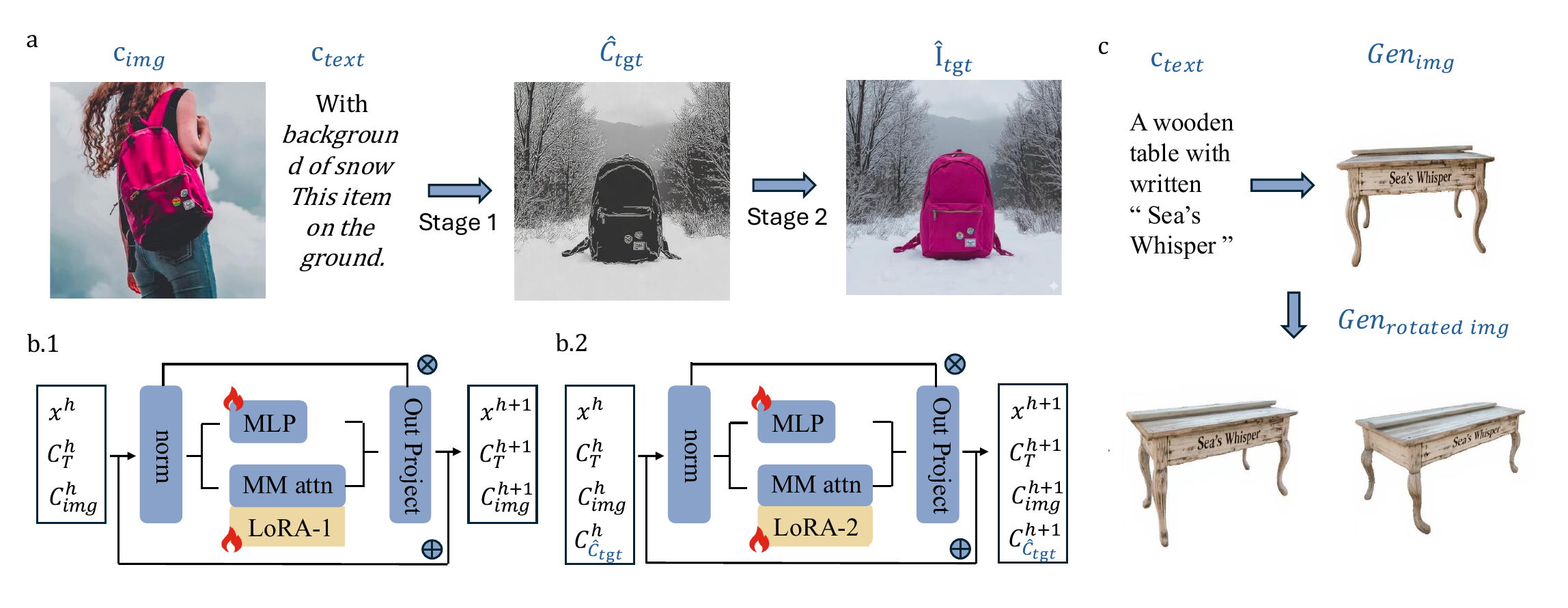}
  \caption{
    \textbf{An overview of our proposed two-stage framework, data pipeline, and network architecture.} 
    \textbf{(a) Two-Stage Inference Pipeline.} Our method decomposes image generation into two distinct stages. Given a source image ($c_{img}$) and a text prompt ($c_{text}$), Stage 1 is dedicated to predicting the target structure, yielding a Canny edge map ($\hat{C}_{tgt}$). This map precisely defines the geometry, pose, and textual layout of the desired output. Stage 2 then takes this structural blueprint, along with the original source image and text prompt, as joint conditions to render the final, full-color image ($\hat{I}_{tgt}$), ensuring that both the subject's appearance and the new structure are faithfully respected.
    \textbf{(b) Unified Network Architecture with Dual LoRAs.} Both generation stages are powered by an identical base architecture (a frozen DiT backbone), but are specialized for their unique tasks via two separate, lightweight LoRA modules. \textbf{(b.1)} The Stage 1 network is conditioned on the source image and text, with its behavior adapted by \textbf{LoRA-1} to focus exclusively on structural prediction. \textbf{(b.2)} The Stage 2 network is conditioned on the source image, text, and the newly generated Canny map. Its behavior is adapted by \textbf{LoRA-2}, which learns to synthesize the final appearance by "coloring in" the provided structure while preserving the identity from the source image. $h$ denotes the $h-$block in our network.
    \textbf{(c) Text-Aware Data Generation Pipeline.} To enable robust text handling, we developed a pipeline to create our `TextingSubject100k` dataset. The process starts with a text-to-image model generating an image ($Gen_{img}$) based on a descriptive prompt ($c_{text}$). Next, to introduce viewpoint variance, we employ Bagel, a 3D-aware novel view synthesizer, to create a rotated version of the object ($Gen_{rotated img}$). Finally, as a critical quality control step, both images are passed through an OCR filter. Only pairs where the text is clearly legible and identical are retained, resulting in a high-quality dataset of corresponding image pairs.
  }
  \label{fig:overview}
\end{figure*}

Our work introduces a novel framework for subject-driven image customization that excels in preserving the intricate identity of a subject, including complex textual elements. The core of our approach is to strategically decompose the complex, direct-to-RGB generation task—a process often prone to identity dilution and entanglement of structure and appearance—into a more manageable, sequential two-stage process. As illustrated in Fig.~\ref{fig:overview}, this "predict-then-render" paradigm first establishes a robust structural blueprint of the target image before proceeding to the final rendering. This decoupling of geometry and appearance not only enhances identity preservation but also constitutes a more efficient and controllable learning paradigm. The entire pipeline is powered by a parameter-efficient network architecture and a specialized dataset meticulously designed for text-aware customization.

\subsection{Two-Stage Generation Process}
\label{sec:two_stage_process}
\begin{figure}[t]
    \centering
    \includegraphics[width=0.9\linewidth]{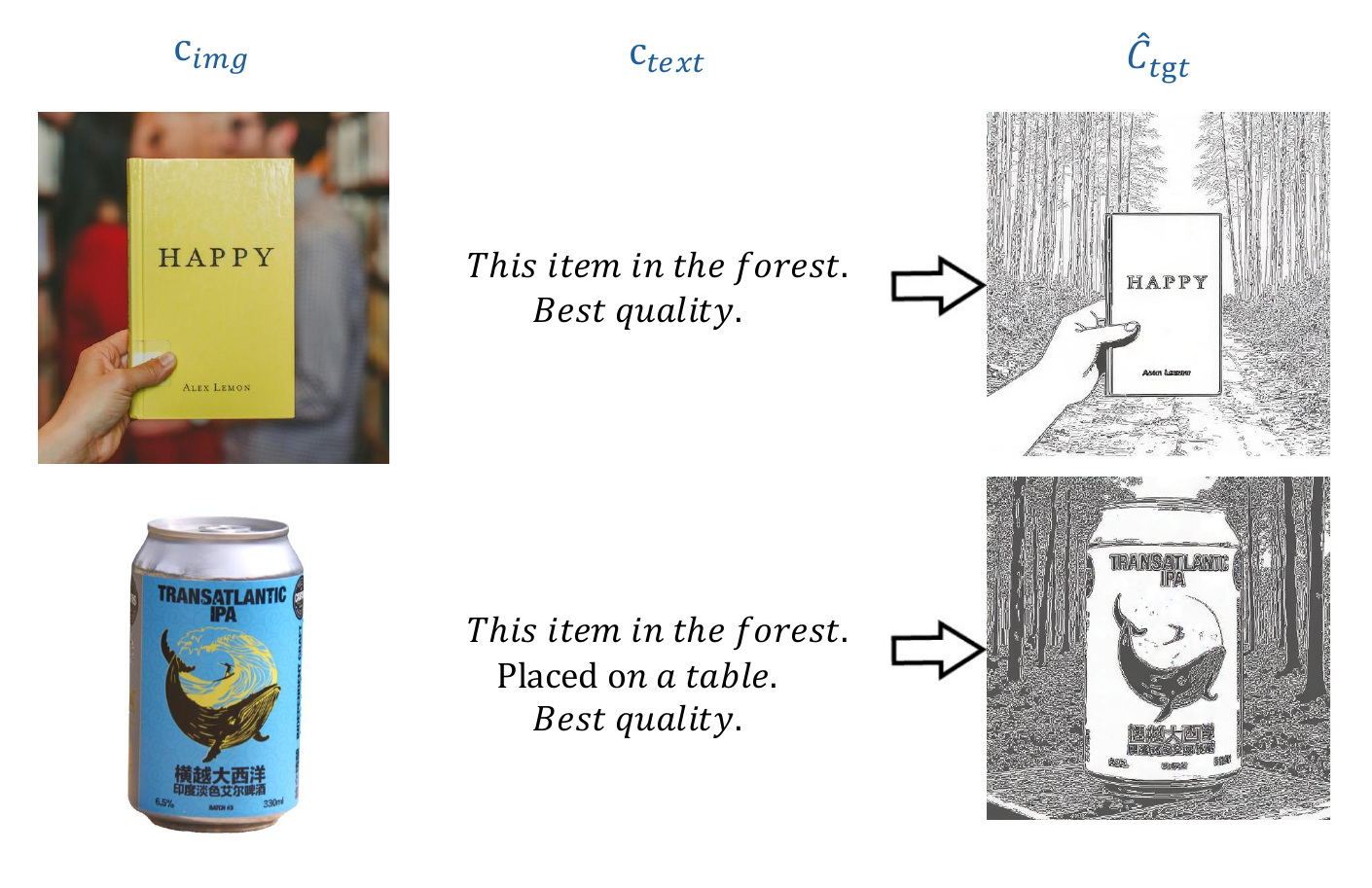}
    \caption{
        \textbf{Demonstration of the text preservation capability of our Stage 1 Canny predictor.} These examples highlight the effectiveness of training with our `TextingSubject100k` dataset. 
    }
    \label{fig:text_preservation}
\end{figure}

Our inference pipeline, visually outlined in Fig.~\ref{fig:overview}a, forms the cornerstone of our methodology. It deliberately avoids the pitfalls of end-to-end generation by breaking the task into two sequential, specialized stages. This process employs two separately trained generator modules, $\mathcal{G}_{\theta_1}$ and $\mathcal{G}_{\theta_2}$, which share an identical underlying architecture but are optimized for their respective tasks using distinct sets of LoRA weights, $\theta_1$ and $\theta_2$.
\paragraph{Stage 1: Target Canny Prediction with $\mathcal{G}_{\theta_1}$.}
The first stage is responsible for establishing the structural foundation of the target image. Its objective is to learn the conditional distribution $p(C_{tgt} | I_{src}, T)$, effectively translating the user's intent into a geometric blueprint. We choose Canny edge maps as our structural representation because they effectively capture high-frequency details like object contours, internal geometry, and, crucially, the precise outlines of text, while abstracting away color and texture information that belongs to the second stage.
The conditioning context for this stage is $c_{stage1} = \{\mathcal{E}_{img}(I_{src}), \mathcal{E}_{text}(T)\}$, where $\mathcal{E}_{img}$ and $\mathcal{E}_{text}$ are pretrained image and text encoders, respectively. The network $\mathcal{G}_{\theta_1}$ is trained using a flow matching objective to predict the target Canny map $C_{tgt}$. To avoid saturating the binary edge target at the image-range boundaries, $C_{tgt}$ is extracted from the ground-truth image, treated as a 3-channel image, and linearly remapped to the range $[0.2, 0.8]$ during training. This preserves the separation between edge and non-edge pixels while leaving margin away from exact 0/1 values, which improves stability in our flow-matching setup. 
Meanwhile, the training data is a curated 80/20 mix of the general-purpose Subject200k dataset and our specialized `TextingSubject100k` dataset (see Sec.~\ref{sec:data_generation}). The inclusion of `TextingSubject100k` is critical; it endows the model with a stronger understanding of typography. As demonstrated in Fig.~\ref{fig:text_preservation}, this allows our model to better preserve complex and stylized text from the source image when generating the target structural map. At inference time, the model produces the predicted Canny map: $\hat{C}_{tgt} = \mathcal{G}_{\theta_1}(z_0, c_{stage1})$.
\paragraph{Stage 2: Canny-Guided Joint Generation with $\mathcal{G}_{\theta_2}$.}
With the structural blueprint $\hat{C}_{tgt}$ firmly established, the second stage undertakes the task of rendering the final, full-color image. The goal here is to learn the distribution $p(I_{tgt} | I_{src}, T, \hat{C}_{tgt})$, effectively "coloring in" the Canny structure while respecting the source subject's identity. We use the second module, $\mathcal{G}_{\theta_2}$, which, while architecturally identical to the first, is trained specifically for this synthesis task.
The conditioning context is expanded to include the powerful structural guidance from Stage 1: $c_{stage2} = \{\mathcal{E}_{img}(I_{src}), \mathcal{E}_{text}(T), \mathcal{E}_{canny}(\hat{C}_{tgt})\}$, where $\mathcal{E}_{canny}$ is an additional lightweight encoder for the predicted Canny map. The Multi-Modal Attention mechanism (detailed in Sec.~\ref{sec:arch_and_training}) now faces a more complex fusion task: it must seamlessly merge information from all three sources. This joint conditioning is what allows the model to resolve potential conflicts and achieve its high fidelity. It simultaneously respects the object's inherent appearance (color, material, texture) from $I_{src}$ and adheres to the desired new structure (pose, composition, text layout) dictated by $\hat{C}_{tgt}$. During training, we use the ground-truth $C_{tgt}$ for stability and to provide a clean learning signal. At inference, the final, customized image is generated as: $\hat{I}_{tgt} = \mathcal{G}_{\theta_2}(z_0, c_{stage2})$.

\vspace{-.6em}
\subsection{Unified Architecture and Training Objective}
\label{sec:arch_and_training}
The generative process for both stages is modeled using a conditional flow matching objective~\cite{lipman2022flow}, which provides a stable and efficient training paradigm well-suited for diffusion-based models. The goal is to learn a continuous-time vector field $v_t(x_t, c)$ that transforms samples from a simple prior distribution $p_0$ (e.g., Gaussian noise) to a target data distribution $p_{data}$, guided by a condition set $c$. The model is trained by minimizing the following loss function:
\begin{equation}
    \mathcal{L}_{FM} = \mathbb{E}_{t, p_t(x_t|x_1), p(x_1)} \left[ || v_t(x_t, c) - (x_1 - x_0) ||_2^2 \right],
\end{equation}
where $t \in [0, 1]$ is the continuous time variable and $p_t(x_t|x_1)$ is a probability path that connects the noise sample to the data sample.
To implement this, we adopt a highly parameter-efficient architectural design, as depicted in Fig.~\ref{fig:overview}b. Our framework is built upon a powerful, pre-trained, and frozen \textbf{FLUX.1-dev}~\cite{flux2024github} backbone. By keeping the backbone frozen, we leverage its robust, general-purpose generative capabilities while significantly reducing computational overhead. The adaptability is introduced via lightweight, trainable LoRA~\cite{hu2022lora} layers, whose parameters are denoted by $\theta$. These LoRA layers are injected into the Multi-Modal Attention (MM-Attn) blocks of the DiT backbone. For a given noisy latent $z_t^l$ and a set of conditioning tokens $\{c_1, ..., c_k\}$, the operation is:
\begin{equation}
    \Delta z_t^l = \text{Proj}_{\text{out}}(\text{MLP}(\text{MM-Attn}(\text{Norm}([z_t^l, c_1, ..., c_k]))))
\end{equation}
This output, $\Delta z_t^l$, then modulates the original latent via a residual connection: $z_t^{l+1} = z_t^l + \Delta z_t^l$. The true elegance of our approach lies in the dual application of this architecture. We train two separate sets of LoRA weights, $\theta_1$ and $\theta_2$, creating two expert modules, $\mathcal{G}_{\theta_1}$ and $\mathcal{G}_{\theta_2}$, from the same base model. This allows for specialized optimization on two different tasks without the need to train two full, multi-billion parameter models.
\subsection{Data Generation for Text-Aware Customization}
\label{sec:data_generation}
A primary obstacle to training generative models for high-fidelity text rendering on customized objects is the profound lack of suitable training data. Existing datasets rarely contain paired images that show the same object with consistent, legible text across different poses or contexts. To overcome this critical bottleneck, we developed a fully automatic data generation pipeline, shown in Fig.~\ref{fig:overview}c, to create a 100k-sample dataset we name `TextingSubject100k`.
The pipeline proceeds in three main steps. First, we leverage a powerful text-to-image model, Seedream 3.0~\cite{gao2025seedream}, to synthesize a diverse set of images, each showing an object with specific text rendered upon it. This provides the initial pool of text-bearing subjects. Second, to introduce the geometric variance necessary for learning robust identity preservation, we employ a 3D-aware novel view synthesizer, Bagel~\cite{deng2025emerging}. For each generated image, Bagel produces new, rotated views of the textualized object, creating image pairs that share the same core identity but differ in perspective.
Third, and most importantly, we implement a stringent quality control stage to ensure the fidelity of the textual information. We apply a robust Optical Character Recognition (OCR) model to both images in every generated pair. A pair is accepted into the final dataset only if the detected text is non-empty, perfectly legible, and identical in both views. This strict filtering protocol discards noisy or inconsistent examples, resulting in a high-quality dataset of corresponding tuples $(I_{src}, I_{tgt}, T)$. Here, $I_{src}$ and $I_{tgt}$ are the consistent image pairs, and $T$ is the generative prompt. This dataset is the crucial ingredient that enables our structure-prediction stage to handle complex textual details with high accuracy.


%% file: sec/4_exps.tex
\section{Experiments}
\label{sec:exps}

\subsection{Setup}

\paragraph{Task and Base Model.}
We evaluate our method on the challenging task of subject-driven image generation. The goal is to synthesize novel images of a specific object, guided by a text prompt, while faithfully preserving its core identity. Our framework is built upon the powerful and efficient FLUX.1-dev~\cite{flux2024github} backbone. We keep its 12.0B parameters frozen during training, ensuring that our approach remains lightweight, computationally efficient.

\paragraph{Implementation Details.}
Our method introduces two lightweight, trainable components: the Canny prediction network (15M parameters) and a minimal control module for the joint generation stage (32M parameters). Both are inspired by the parameter-efficient paradigm of OminiControl~\cite{tan2024ominicontrol}. As shown in Tab.~\ref{tab:params}, the total parametric overhead is a mere \textasciitilde0.5\% of the base model, demonstrating remarkable efficiency. We use the Prodigy optimizer~\cite{mishchenko2023prodigy} with a batch size of 32 and a constant learning rate of 1e-4. All models are trained on 8 NVIDIA H100 GPUs for approximately 2 days.

\begin{table}[h]
\centering
\caption{Parameter comparison of different conditioning methods on FLUX.1-dev. Our method achieves strong control with minimal parametric overhead.}
\label{tab:params}
\begin{tabular}{l|c|r|r}
\Xhline{1.2pt} 
Methods & Base model & Parameters & Ratio \\
\hline\hline 
IP-Adapter~\cite{ye2023ip} & \multirow{4}{*}{\makecell{FLUX.1-dev \\ (12.0B)}} & 45M & \textasciitilde0.4\% \\
SSR-Encoder~\cite{zhang2024ssr} & & 120M & \textasciitilde1.0\% \\
OminiControl~\cite{tan2024ominicontrol} & & 15M & \textasciitilde0.1\% \\
\cline{1-1} \cline{3-4} 
\rowcolor{gray!20} 
Ours & & \textbf{15M+32M} & \textbf{\textasciitilde0.5\%} \\
\Xhline{1.2pt} 
\end{tabular}
\end{table}
\vspace{-1.em}
\paragraph{Training Data.}
Our training regimen relies on a strategic combination of two datasets: the public, general-purpose Subject200k dataset~\cite{tan2024ominicontrol} and our proposed, specialized TextingSubject100k dataset. This mixed-data strategy is crucial, as it allows the model to learn robust, general object consistency from a broad distribution while simultaneously acquiring the specialized ability to render high-fidelity text and patterns from our targeted data.

\paragraph{Evaluation Protocol.}
To ensure a rigorous and reproducible assessment, we adopt a robust, automated evaluation protocol using the advanced vision-language model, \textbf{GPT-4.1} (`gpt-4.1-2025-04-14`). This approach moves beyond the subjectivity and high cost of human studies. For each subject-prompt pair from the standard DreamBooth benchmark~\cite{ruiz2022dreambooth}, the VLM assesses the generated image against the source image, providing fine-grained scores (from 0 to 10) across three key dimensions:
\textbf{Subject Consistency (SC):} Measures the preservation of identity, including shape, color, and fine details. This is our primary metric for fidelity. \textbf{Prompt Adherence (PA):} Assesses how well the generated image follows the textual instructions regarding background, action, and style. \textbf{Perceptual Quality (PQ):} Evaluates the overall visual quality, considering both image naturalness and the absence of artifacts. The final score is the minimum of the two sub-scores, ensuring a strict quality standard.
This multi-faceted evaluation provides a holistic view of model performance, capturing the delicate balance between fidelity and editability. For completeness, we also report standard automated metrics: Fréchet Inception Distance (FID)~\cite{heusel2017gans} and CLIP Score~\cite{radford2021learning}.
No single automatic metric fully captures subject fidelity, so these scores should be interpreted jointly with the qualitative comparisons. Broader perceptual metrics such as DINO-based similarity or DreamSim would be valuable extensions for future work.
\begin{table}[t!]
\centering
\caption{Main results on the DreamBooth benchmark. Our method demonstrates superior performance across all metrics, especially in subject consistency.}
\label{tab:main_results}
\resizebox{\columnwidth}{!}{%
\begin{tabular}{@{}l|ccc|cc@{}}
\toprule
\multirow{2}{*}{Method} & \multicolumn{3}{c|}{VLM-based Metrics $\uparrow$} & \multicolumn{2}{c}{Auto Metrics} \\
\cmidrule(l){2-6} 
& SC & PA & PQ & FID $\downarrow$ & CLIP $\uparrow$ \\
\midrule
IP-Adapter~\cite{ye2023ip} & 4.05 & 6.12 & 5.01 & 144.51 & 0.607 \\
SSR-Encoder~\cite{zhang2024ssr} & 5.12 & 6.33 & 5.43 & 128.12 & 0.680 \\
OminiControl~\cite{tan2024ominicontrol} & 6.05 & 6.88 & 6.45 & 102.12 & 0.767 \\
\textbf{Ours} & \textbf{7.05} & \textbf{7.15} & \textbf{6.93} & \textbf{78.12} & \textbf{0.820} \\
\bottomrule
\end{tabular}%
}
\end{table}
\vspace{-1.2em}
\paragraph{Baselines.}
We compare our method against three strong and representative baselines: IP-Adapter~\cite{ye2023ip}, a seminal work in adapter-based tuning; SSR-Encoder~\cite{zhang2024ssr}, a more recent encoder-based approach; and a modified OminiControl~\cite{tan2024ominicontrol} adapted for direct source image conditioning. For a fair and rigorous comparison, all baselines are trained from scratch by us using the exact same mixed dataset, training schedule, and base model.

\subsection{Main Quantitative and Qualitative Results}

\paragraph{Quantitative Comparison.}
Tab.~\ref{tab:main_results} presents the comprehensive results on the DreamBooth benchmark. Our method consistently and significantly outperforms all baselines across every VLM-based metric and standard automated metric. The most noteworthy result is in \textbf{Subject Consistency (SC)}, where our method achieves a score of 7.05. This represents a 16.5\% relative improvement over the next best baseline (OminiControl at 6.05), a substantial margin that underscores the effectiveness of our approach. This large gap directly validates our core hypothesis: by decoupling structure and appearance via an intermediate Canny prediction, our two-stage approach can preserve high-frequency identity details far more effectively than methods that attempt a direct, entangled, end-to-end generation. While excelling in consistency, our method also achieves the highest scores in Prompt Adherence (PA) and Perceptual Quality (PQ), indicating that our focus on structure does not come at the cost of editability or visual fidelity. The superior FID and CLIP scores further corroborate these findings, reflecting a better overall distribution match and text-image alignment.
\paragraph{Qualitative Analysis.}
As vividly illustrated in our teaser (Fig.~\ref{fig:teaser}), the quantitative superiority of our method translates into visually compelling and often dramatically better results. In challenging scenarios involving subjects with intricate patterns or text—such as the beverage can with a detailed logo or the patterned backpack—our method renders these details with remarkable clarity and correctness. In contrast, baseline methods often produce blurred, distorted, or entirely incorrect text and patterns, demonstrating their fundamental failure to capture and propagate high-frequency information. Furthermore, when subjected to significant stylistic or structural edits, such as turning a specific pet into a "roman statue," our method successfully preserves the unique facial structure and identity of the pet within the new style. Baselines, however, tend to drift towards a generic statue appearance, losing the subject's core identity in the process. These qualitative examples strongly support our claim that explicitly modeling structure is the key to achieving true, robust subject fidelity.

\subsection{Validating the Efficacy of the Two-Stage Learning Paradigm}

\paragraph{Evaluation of Canny Generation.}
To validate the foundation of our framework, we first isolate and evaluate the Canny prediction stage ($\mathcal{G}_{\theta_1}$). The goal is to test its ability to act as a high-fidelity "structure carrier." We task the model with reconstructing the source Canny map given a neutral prompt ("this item, without any change"). As shown in Tab.~\ref{tab:canny_recon}, the model achieves an exceptionally high PSNR of 35.12 and SSIM of 0.94. This near-perfect reconstruction demonstrates that the first stage does not act as an information bottleneck for structural details. Instead, it can accurately capture and reconstruct the essential structure of the source subject, forming a solid and reliable foundation for the subsequent generation stage.

\begin{table}[h]
\centering
\caption{Quantitative analysis of the Canny prediction stage's ability to reconstruct the source structure given a neutral prompt.}
\label{tab:canny_recon}
\begin{tabular}{@{}lcc@{}}
\toprule
Method & PSNR $\uparrow$ & SSIM $\uparrow$ \\
\midrule
\textbf{Ours (Canny Predictor)} & \textbf{35.12} & \textbf{0.94} \\
\bottomrule
\end{tabular}
\end{table}

\paragraph{Knowledge Distillation Experiment.}
Using synthetic supervision from Seedream~4~\cite{seedream2025seedream}, we compare standard FLUX.Kontext~\cite{batifol2025flux} fine-tuning against our two-stage method ("via ISP"). As shown in Tab.~\ref{tab:distillation}, ISP yields higher SC and PA scores than the fine-tuned baseline, suggesting that decomposition is a useful learning route in this setting.

\begin{table}[h]
\centering
\caption{Evaluation of models trained on a dataset distilled from Seedream4. Our two-stage method ("via ISP") leverages the distilled data more effectively than standard fine-tuning.}
\label{tab:distillation}
\begin{tabular}{@{}l|ccc@{}}
\toprule
Method & SC $\uparrow$ & PA $\uparrow$ & PQ $\uparrow$ \\
\midrule
Seedream4 (Teacher)~\cite{seedream2025seedream} & 9.75 & 9.85 & 7.72 \\
\hline
FLUX.Kontext~\cite{batifol2025flux} & 8.32 & 9.09 & 6.88 \\
FLUX.Kontext-finetuned & 8.46 & 9.21 & 7.32 \\
\textbf{Ours (via ISP)} & \textbf{8.86} & \textbf{9.40} & \textbf{7.38} \\
\bottomrule
\end{tabular}
\end{table}

\subsection{Ablation Studies}
\label{sec:ablation}

\paragraph{Effect of LoRA Rank.}
The capacity of our trainable modules is controlled by the LoRA rank. As shown in Tab.~\ref{tab:ablation_lora}, we observe a clear trend: increasing the rank from 2 to 16 yields consistent improvements across all metrics. The performance jump from rank 2 to 8, particularly in SC, indicates that capturing the complex mapping from image+text to structure requires significant model capacity. The performance begins to saturate at rank 16, which offers the best balance of performance and parameter count before hitting diminishing returns. This justifies our choice of rank 16 for all main experiments.
\vspace{-1.2em}
\begin{table}[h]
\centering
\caption{Ablation on LoRA rank. A rank of 16 performs the best.}
\label{tab:ablation_lora}
\begin{tabular}{c|ccc}
\toprule
LoRA Rank & SC $\uparrow$ & PA $\uparrow$ & PQ $\uparrow$ \\
\midrule
2         & 6.48           & 7.29           & 6.52               \\
4         & 6.60           & 7.13           & 6.64               \\
8         & 6.91           & 7.23           & 6.82               \\
\textbf{16} & \textbf{7.05}  & 7.15          & \textbf{6.93}      \\
\bottomrule
\end{tabular}
\end{table}
\paragraph{Effect of Data Composition.}
Our mixed-data strategy is critical for achieving a well-rounded model. Tab.~\ref{tab:ablation_data} reveals an important trade-off between generalization and specialization. Training solely on the general-purpose Subject200k yields a competent but generic model that fails on fine details (lower SC). Conversely, training only on our specialized `TextingSubject100k` dataset creates a "specialist" model that excels at prompt following (highest PA) but suffers from reduced generalization on non-textual objects. The performance drop at a 50/50 split is particularly insightful, possibly indicating a "gradient conflict" where the model struggles to reconcile two equally strong but different data distributions. The 80/20 mixture provides the best of both worlds, using the larger dataset to establish a robust generative prior, which is then effectively "sharpened" on the specialized text data. This validates our data strategy.

\begin{table}[h]
\centering
\caption{Ablation on training data composition. The 80/20 split provides the best balance between generalization.}
\label{tab:ablation_data}
\begin{tabular}{l|ccc}
\toprule
Data Composition & SC $\uparrow$ & PA $\uparrow$ & PQ $\uparrow$ \\
\midrule
100\% Subject200k & 6.76 & 7.22 & 6.85 \\
100\% TextingSubject100k & 6.85 & \textbf{7.42} & 6.90 \\
50\% S200k + 50\% Ours & 6.90 & 7.00 & 6.75 \\
\textbf{80\% S200k + 20\% Ours} & \textbf{7.05} & 7.15 & \textbf{6.93} \\
\bottomrule
\end{tabular}
\end{table}

\paragraph{Contribution of TextingSubject100k.}
To specifically quantify the impact of our `TextingSubject100k` dataset, we evaluate on a challenging text-rendering benchmark, TextBench. As shown in Tab.~\ref{tab:ablation_text}, training with our dataset boosts OCR accuracy from a mediocre 65.2\% to a remarkable 85.7\%. This is not a minor improvement; it represents a qualitative leap from "mostly illegible" to "highly accurate" rendering. It confirms that for highly specific and structured sub-tasks like typography, general-purpose datasets are insufficient, and targeted, high-quality data collection, as enabled by our automated pipeline, is essential for breakthrough performance.
\vspace{-1.em}
\begin{table}[h]
\centering
\caption{Focused evaluation on text rendering using OCR accuracy (\%) on the TextBench benchmark. Our specialized dataset is crucial for high-fidelity text generation.}
\label{tab:ablation_text}
\begin{tabular}{l|c}
\toprule
Method & OCR Accuracy $\uparrow$ \\
\midrule
Ours (trained on Subject200k only) & 65.2\% \\
\textbf{Ours (trained w/ TextingSubject100k)} & \textbf{85.7\%} \\
\bottomrule
\end{tabular}
\end{table}
\vspace{-.4em}

%% file: sec/5_conclu.tex
\section{Conclusion}
\label{sec:conclusion}

We address the challenge of preserving high-fidelity details in subject-driven image generation with a two-stage framework that decouples structure from appearance. By first predicting a Canny map and then rendering the final image conditioned on that structure, the method provides a more structured learning route; we further support text-aware generation with the automatically constructed `TextingSubject100k` dataset.

Our experiments, evaluated by a GPT-4.1-based protocol, show strong gains over the selected baselines in subject consistency, prompt adherence, and perceptual quality. The knowledge distillation experiment further suggests that the proposed decomposition can make effective use of high-quality synthetic supervision in the setting we study.

Despite its strong performance, our method has limitations. The sequential nature of the two-stage process introduces additional inference latency compared to single-pass models, and the final output still depends on the quality of the initial Canny prediction. More importantly, our current validation is concentrated on FLUX.1-dev and a single FLUX.Kontext adaptation rather than a wider family of larger and more recent backbones. We therefore view the present work less as a final answer and more as a methodological route: it suggests that explicit structural decomposition is a promising technical path for future subject-driven generation research. Exploring broader backbone coverage, richer perceptual metrics, and alternative structural guides such as depth or segmentation remains important future work.


%% file: sec/X_supp.tex
\clearpage
\appendix
\onecolumn 

\begin{center}
{\LARGE\bfseries Decomposing Subject-Driven Image Generation \\
via Intermediate Structural Prediction \par}
\vspace{0.8em}
{\large Supplementary Material \par}
\vspace{1.0em}
Hanzhong Guo \qquad \quad Yizhou Yu \par
\vspace{0.5em}
School of Computing and Data Science, The University of Hong Kong  \par
\vspace{0.3em}
\texttt{hanzhong@connect.hku.hk, yizhouy@acm.org} \par
\end{center}
\vspace{1.2em}

\section{Additional Qualitative Results}
\label{sec:appendix_qualitative}

This section provides additional qualitative results to further demonstrate the effectiveness of our method. We first compare our approach against existing baseline models and then showcase more generation results for objects both with and without text.

\begin{figure*}[h!]
    \centering
    \includegraphics[width=0.9\textwidth]{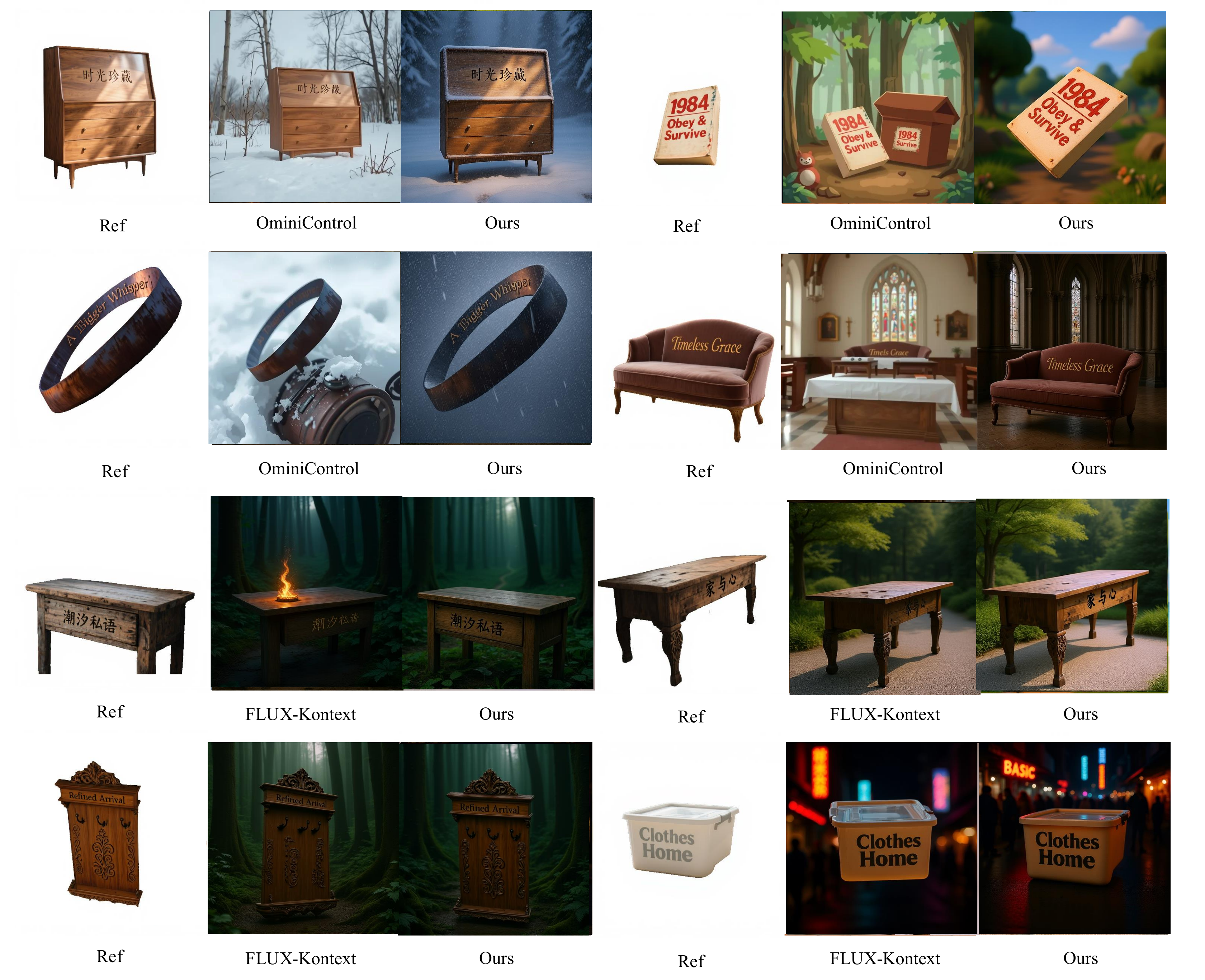}
    \caption{
        \textbf{Qualitative comparison with baseline methods.}
        Each example shows the reference image (Ref) alongside the outputs from baseline models (OminiControl, FLUX-Kontext) and our method.
        Compared to the baselines, our method demonstrates superior performance in generating images that are more consistent with the text prompt while more faithfully preserving the subject's identity and intricate textual details.
    }
    \label{fig:appendix_comparison}
\end{figure*}

\begin{figure*}[h!]
    \centering
    \includegraphics[width=\textwidth]{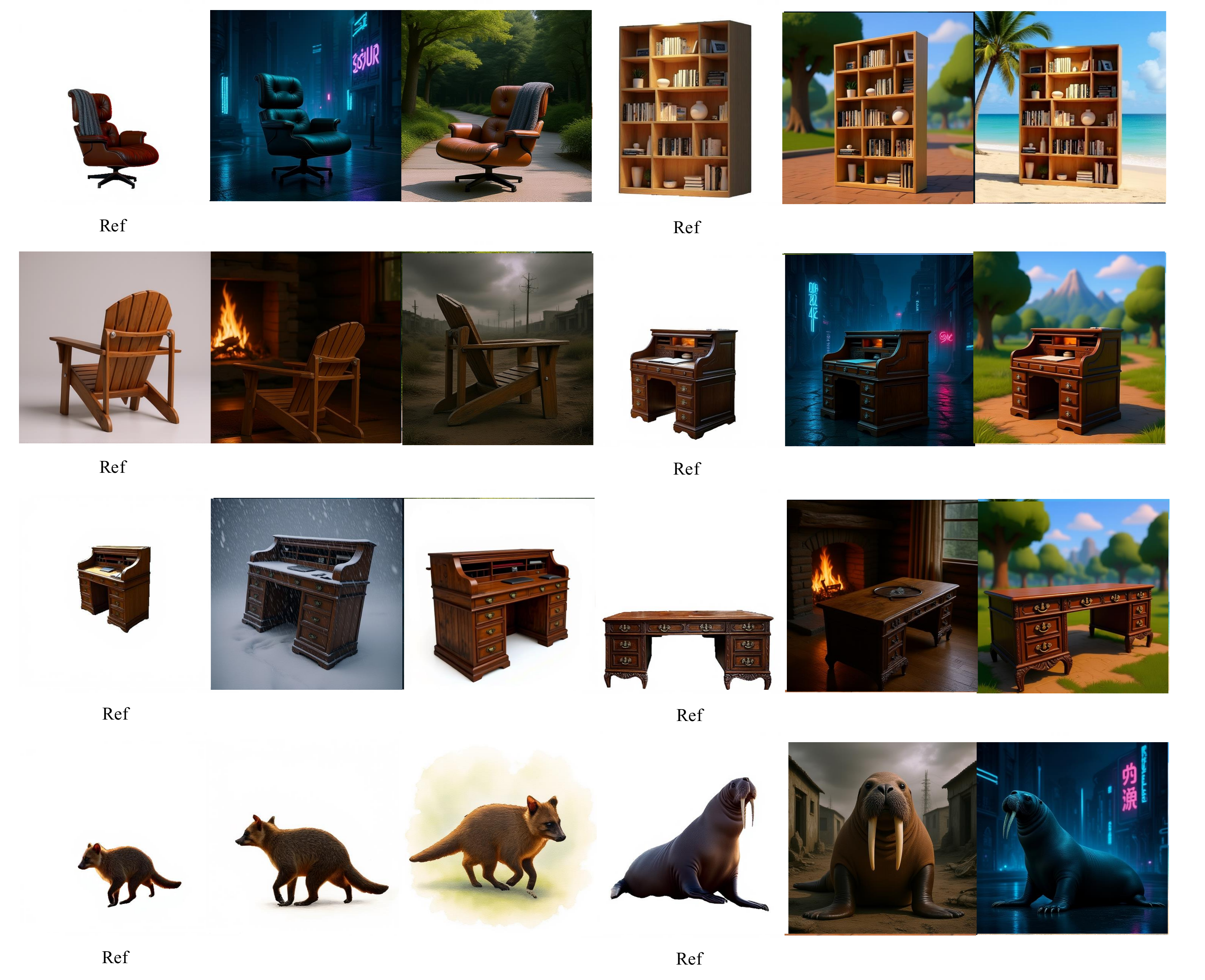}
    \caption{
        \textbf{Additional qualitative results on text-free objects.}
        Each group displays a reference image (Ref) and two different scenes generated by our model.
        These examples showcase the model's ability to robustly place the source object into diverse new environments and styles while maintaining its core identity features.
    }
    \label{fig:appendix_qualitative_simple}
\end{figure*}

\begin{figure*}[h!]
    \centering
    \includegraphics[width=\textwidth]{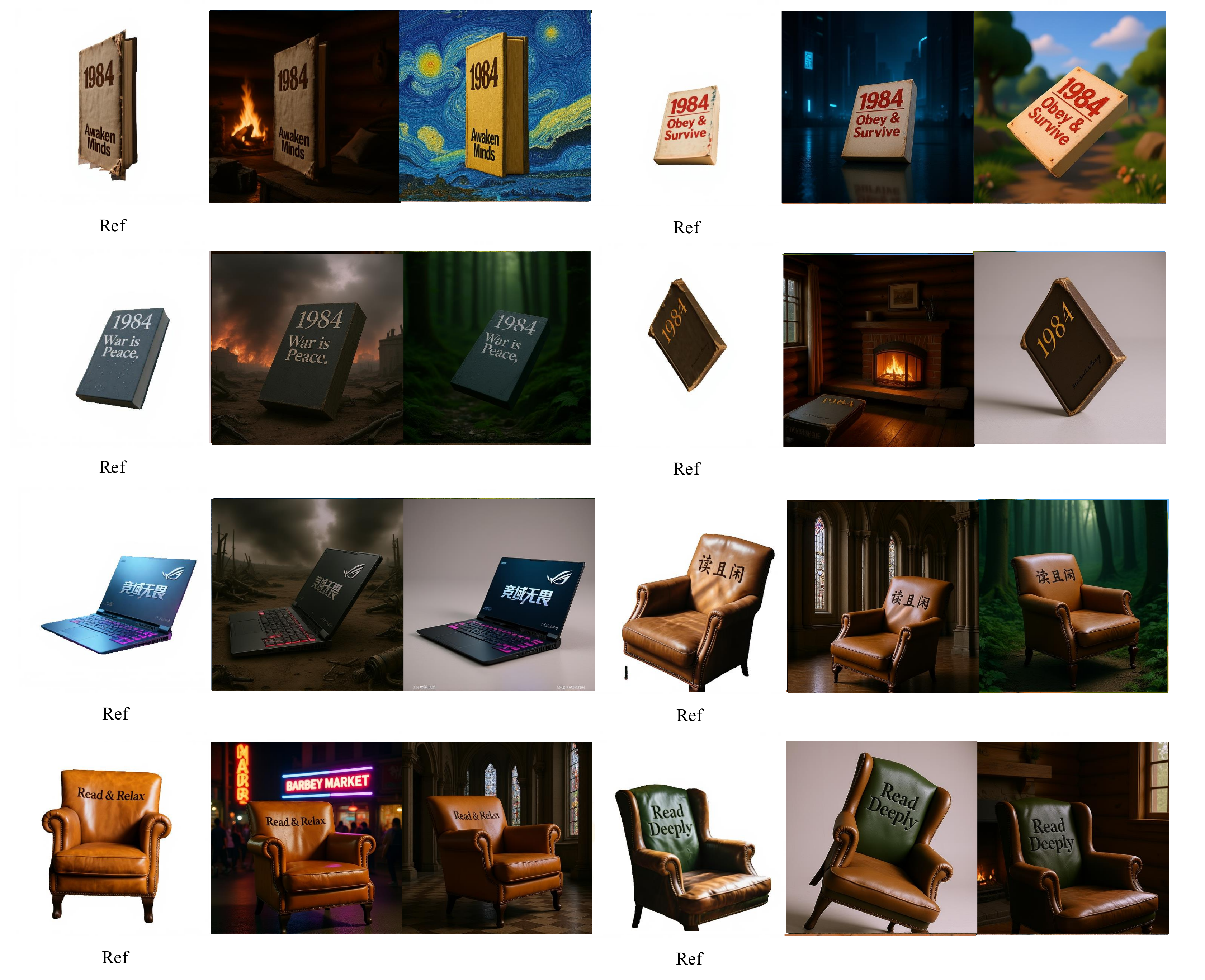}
    \caption{
        \textbf{Additional qualitative results on objects with text.}
        Each group shows a reference image (Ref) with text and two outputs generated in new scenes.
        Our method successfully preserves the legibility, style, and content of the text on the object even under significant changes in background and lighting, a key contribution of our work.
    }
    \label{fig:appendix_qualitative_text}
\end{figure*}

\begin{figure*}[h!]
    \centering
    \includegraphics[width=\textwidth]{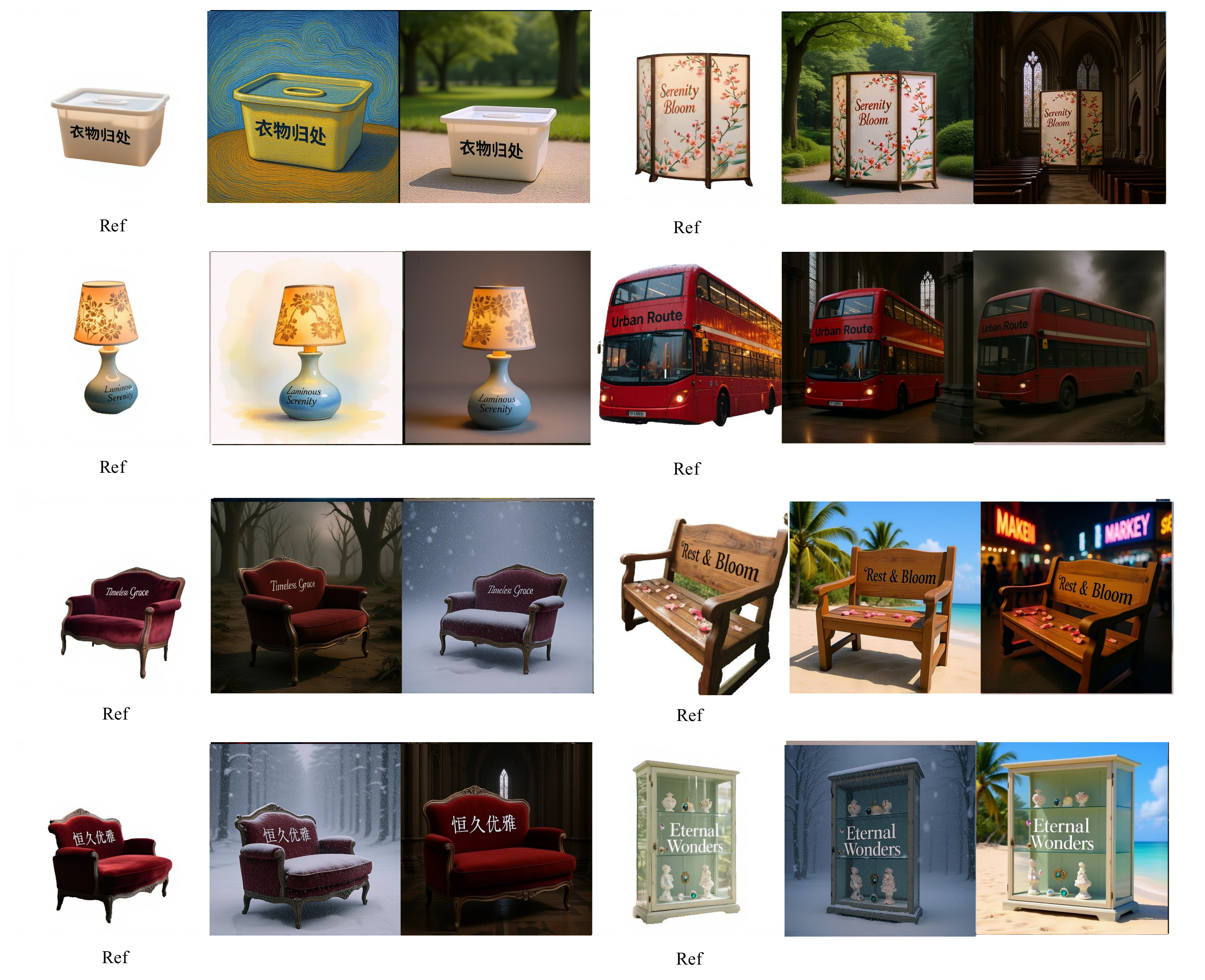}
    \caption{
        \textbf{Additional qualitative results on objects with text.}
        Each group shows a reference image (Ref) with text and two outputs generated in new scenes.
        Our method successfully preserves the legibility, style, and content of the text on the object even under significant changes in background and lighting, a key contribution of our work.
    }
    \label{fig:appendix_qualitative_text2}
\end{figure*}

\clearpage
\section{Dataset Examples}
\label{sec:appendix_dataset}

This section provides examples from the datasets used to train our model. We augmented the general-purpose `Subject200k` dataset and constructed the specialized `TextingSubject100k` dataset to handle text rendering tasks.

\begin{figure*}[h!]
    \centering
    \includegraphics[width=\textwidth]{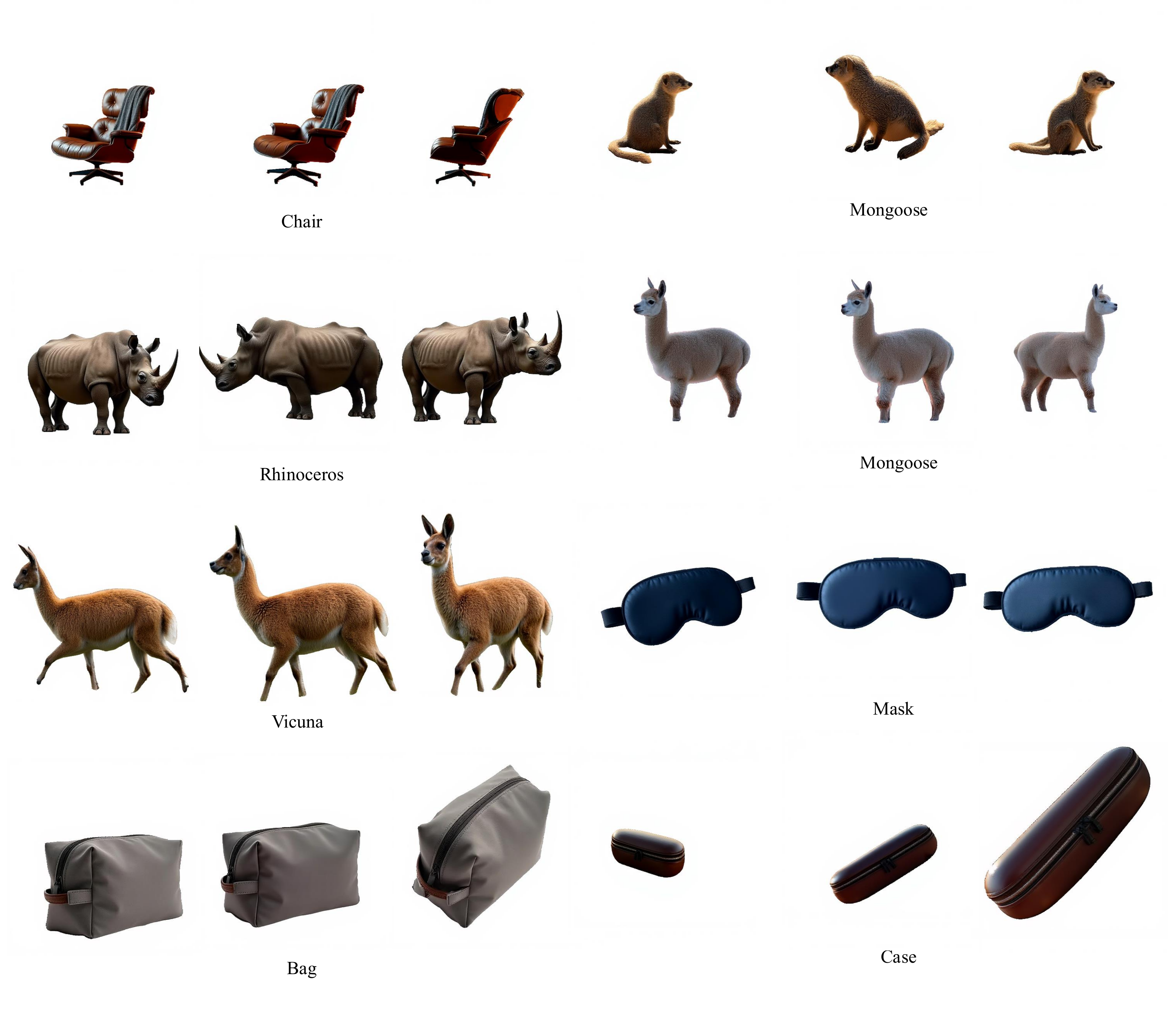}
    \caption{
        \textbf{Examples from the augmented `Subject200k` dataset.}
        These image triplets are created by taking a source image and applying Bagel to generate novel, rotated views.
        This data augmentation strategy helps the model learn a robust representation of object identity across different viewpoints.
    }
    \label{fig:appendix_data_subject200k}
\end{figure*}

\begin{figure*}[h!]
    \centering
    \includegraphics[width=\textwidth]{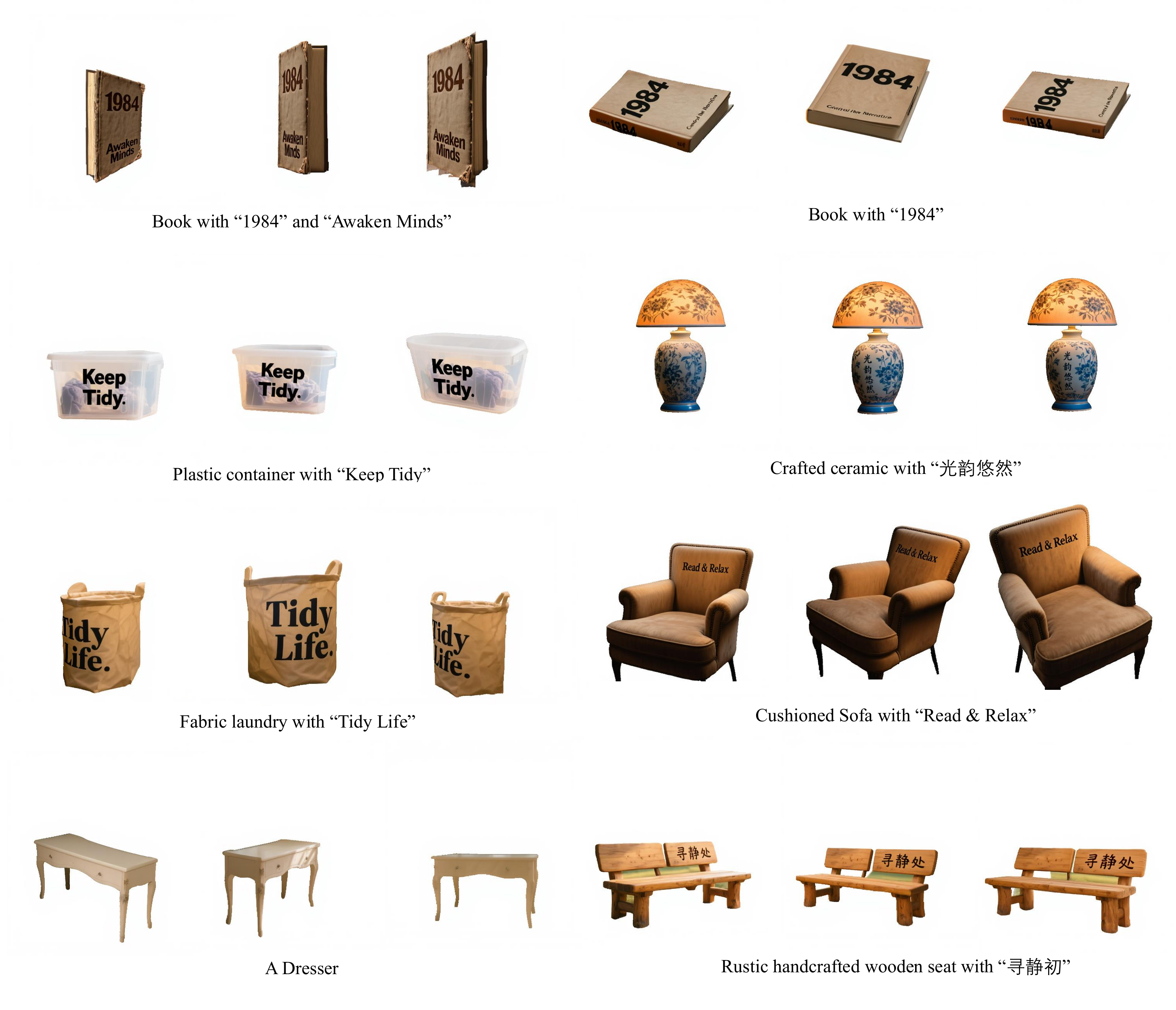}
    \caption{
        \textbf{Examples from our custom `TextingSubject100k` dataset.}
        Each triplet shows an object with text, with its views rotated by Bagel.
        We employ a strict OCR filtering process to ensure the text remains consistent and legible across views, providing high-quality training data for our text-aware model.
    }
    \label{fig:appendix_data_textingsubject}
\end{figure*}
